\documentclass[conference]{IEEEtran}
\IEEEoverridecommandlockouts
\usepackage{cite}
\usepackage{amsmath,amssymb,amsfonts}
\usepackage{algorithmic}
\usepackage{graphicx}
\usepackage{textcomp}
\usepackage{xcolor}
\def\BibTeX{{\rm B\kern-.05em{\sc i\kern-.025em b}\kern-.08em
    T\kern-.1667em\lower.7ex\hbox{E}\kern-.125emX}}
    
\usepackage{amsfonts}

\newcommand\norm[1]{\left\lVert#1\right\rVert}

\usepackage{tabularx}
    \newcolumntype{L}{>{\raggedright\arraybackslash}X}
    \newcolumntype{C}{>{\raggedright\arraybackslash}Z}
\newcommand{\heading}[1]{\multicolumn{1}{|c|}{#1}}

\usepackage{siunitx}

\usepackage{algorithm}

\begin{document}

\title{Unified Multi-Domain Learning and Data Imputation using Adversarial Autoencoder}

\author{\IEEEauthorblockN{Andre Mendes}
\IEEEauthorblockA{\textit{New York University} \\
New York, USA \\
andre.mendes@nyu.edu}
\and
\IEEEauthorblockN{Julian Togelius}
\IEEEauthorblockA{\textit{New York University} \\
New York, USA \\
julian.togelius@nyu.edu}
\and
\IEEEauthorblockN{Leandro dos Santos Coelho}
\IEEEauthorblockA{\textit{Pontifical Catholic University of Parana} \\
\textit{Federal University of Parana}\\
Curitiba, Brazil \\
leandro.coelho@pucpr.br}
}

\maketitle

\begin{abstract}
We present a novel framework that can combine multi-domain learning (MDL), data imputation (DI) and multi-task learning (MTL) to improve performance for classification and regression tasks in different domains. The core of our method is an adversarial autoencoder that can: (1) learn to produce domain-invariant embeddings to reduce the difference between domains; (2) learn the data distribution for each domain and correctly perform data imputation on missing data. For MDL, we use the Maximum Mean Discrepancy (MMD) measure to align the domain distributions. For DI, we use an adversarial approach where a generator fill in information for missing data and a discriminator tries to distinguish between real and imputed values. Finally, using the universal feature representation in the embeddings, we train a classifier using MTL that given input from any domain, can predict labels for all domains. We demonstrate the superior performance of our approach compared to other state-of-art methods in three distinct settings, DG-DI in image recognition with unstructured data, MTL-DI in grade estimation with structured data and MDMTL-DI in a selection process using mixed data.
\end{abstract}

\begin{IEEEkeywords}
multi-task, multi-domain, data imputation
\end{IEEEkeywords}

\section{Introduction}
Many real-world problems in machine learning (ML) require enough labeled data for training classifiers to address a task at hand. Whenever this data is either not available, or hard to collect, an alternative is to use available datasets that are related to the task. However, although related, these other tasks often belong to a different domain and training classifiers on them without addressing their differences result in sub-optimal performance. To address this case, multi-domain learning (MDL)~\cite{joshi2012multi} approaches have been proposed to reduce the gap among domains and create more general classifiers.

MDL has different applications, such as Domain Generalization (DG)~\cite{muandet2013domain} when the goal is to use source datasets to learn a classifier for an unknown target domain and Domain Adaptation (DA)~\cite{pan2009survey} when the target domain is known but the labels are not available or their sample size is small. In this work, we focus on a case of MDL where datasets and labels for different domains are available. However, all datasets have relatively small sample sizes. Therefore, all datasets are used as source and target and the knowledge has to be shared in all directions. We propose a framework based on an autoencoder~\cite{chen2016variational} that can produce domain-invariant encoded spaces. We adopt the maximum mean discrepancy (MMD)~\cite{gretton2012kernel} to measure differences between generated encoded spaces, and we reduce MMD during training.

Another important challenge in ML is dealing with incomplete datasets where a significant part of the data is missing. One solution would be to remove the damaged samples, but this is often impractical because of the reduction in sample size. To address this case, Data Imputation (DI)~\cite{buuren2010mice,stekhoven2011missforest,chen2016variational} methods try to learn the data distribution and generate values to fill in for missing data. In our approach, we use an adversarial strategy to incorporate DI in our autoencoder, combining MDL and DI to jointly address missing data and domain differences.

Additionally to this unsupervised approach, we also proposed a supervised method to incorporate the labels available in the training process. To accomplish that, we adapt a multi-task learning (MTL)~\cite{zhang2017survey,caruana1997multitask} framework in which a single classifier is trained to make predictions for all tasks/domains. By learning the tasks together, the classifier is exposed to more samples in different domains, allowing it to improve generalization and perform better than classifiers trained individually. 

Therefore, the main contributions of this work are:
\begin{itemize}
    \item We present an adversarial autoencoder to perform DI and MDL in the same training process. The process is unsupervised and the encoded space and features generated can be used in different downstream tasks.
    \item We present a supervised framework that combines the autoencoder with an MTL approach to train a classifier for all domains/tasks together.
    \item We address an important case of MDL where there is no significant source/target dataset and all datasets have to share knowledge among them.
    \item By performing experiments in different settings, we show that our method can be applied to DG, MTL, and MDL using the same structure. We also show empirically that it outperforms other state-of-art methods in all settings.
\end{itemize}

This paper is organized as follows: Section~\ref{sec:related} shows the related work. Section~\ref{sec:problem_statement} introduces our notation notation and problem statement. Section~\ref{sec:method} explains the method and Section~\ref{sec:experiments} describes the experiments for validation. Section ~\ref{sec:results} presents the results and a conclusion and future work are discussed in Section~\ref{sec:conclusion}.

\section{Related Work}
\label{sec:related}
Here we describe related work in MDL, DI and MTL.
\subsection{Multi-Domain Learning (MDL)}
While traditional ML methods train classifiers to generalize well for a specific domain, MDL trains classifiers that can predict outputs across domains with different data distribution~\cite{joshi2012multi}. For example,~\cite{yang2014unified} uses metadata (semantic descriptors), to explore information on task and domain relatedness. They construct a two-sided network to learn representations from the original input vector and the metadata by minimizing the empirical risk for all domains/tasks. MDL has vast applicability in different contexts~\cite{nam2016learning,yang2014unified}.

DG uses data from seen source domains to create models that can generalize well to unseen target domains. The work in~\cite{muandet2013domain} uses a domain-invariant component analysis method with a kernel-based optimization algorithm to minimize the dissimilarity across domains. Other approaches, such as~\cite{motiian2017unified}, perform DG and DA using a contrastive loss to guide samples from the same class being embedded nearby in latent space across data sources. Similar to our method, the work in~\cite{ghifary2015domain} uses an autoencoder, expanding it to a multi-task approach to extract invariant features that are robust to domain variations. In~\cite{li2018domain}, adversarial autoencoders are extended to DG by imposing MMD~\cite{gretton2012kernel} metric to align multi-domain distributions.

DA~\cite{pan2009survey} is similar to DG, however, in this case the target data is known. DA methods attempt to minimize the shift between source and target distributions in input space~\cite{hoffman2017cycada}, feature space~\cite{ganin2016domain,liu2019transferable}, or output space~\cite{luo2019taking}, using MMD~\cite{sejdinovic2013equivalence} or adversarial learning (GAN)~\cite{goodfellow2014generative}. DA has been applied to computer vision~\cite{liu2019transferable,hoffman2017cycada}, structured data~\cite{long2013adaptation} and text~\cite{liu2019transferable}.
\subsection{Data Imputation (DI)}
Methods for DI range from simple column average to complex imputations, and they can be categorized as discriminative or generative. Discriminative methods include MICE~\cite{buuren2010mice} and matrix completion~\cite{mazumder2010spectral}. Generative methods include algorithms based on Expectation Maximization~\cite{garcia2010pattern} and on deep learning (DL)~\cite{nowozin2016f,arjovsky2017wasserstein,chen2016variational}. 

DL methods can capture the latent structure of high-dimensional data by efficiently emulating complex distributions~\cite{kingma2013auto}. Many of the algorithms proposed in this setting use autoencoder based structures~\cite{tomczak2018vae,chen2016variational} or adversarial generative~\cite{goodfellow2014generative} based methods ~\cite{nowozin2016f,arjovsky2017wasserstein}. While most of these approaches focus on computer vision tasks, the work in~\cite{nazabal2018handling} focuses on solving DI for arbitrary datasets, including tabular data, which are often incomplete and heterogeneous.

In our approach, we deal with structured and unstructured data and our method is similar to~\cite{yoon2018gain}, which explores the GAN approach to create: a generator to accurately impute missing data; a discriminator to distinguish between observed and imputed components; and a hint mechanism to help generate samples according to the true underlying data distribution.
\subsection{Multi-Task Learning (MTL)}
The goal in MTL~\cite{zhang2017survey,caruana1997multitask} is to learn multiple related tasks simultaneously so that knowledge obtained from each task can be re-used by the others. This can increase the sample size for each task, making MTL beneficial for tasks with small training sample. MTL has been applied in different approaches including neural nets (NN)~\cite{caruana1997multitask}, kernel methods~\cite{kumar2012learning}, and deep neural nets (DNN)~\cite{ruder2017overview}. 

Regularization is important in MTL since it controls the relationship among tasks and features and prevents overfitting. The work in~\cite{zhou2011malsar} presents an overview of regularization in MTL, such as group-sparse based~\cite{chen2011integrating} and robust feature based~\cite{kumar2012learning} regularization. Other methods have explored tensor factorization~\cite{yang2016deep} and evolutionary algorithms~\cite{liang2018evolutionary} to find optimal architectures for MTL networks.
\section{Problem Statement}
\label{sec:problem_statement}
Let's define a domain $\mathcal{D}$ with feature space $\mathcal{X}$ and a marginal probability distribution $P(\mathbf{X})$. Each $\mathbf{x}_i \in \mathbb{R}^d$ is the \textit{i}-th feature vector (sample), $n$ is the number of samples and $d$ is the number of dimensions in $\mathbf{X} \in \mathbb{R}^{n{\times}d}$. The \textit{i}-th one-hot encoding label sample is represented by $\mathbf{y}_i \in \mathbb{R}^{d_y}$ with $d_y$ as the number of classes. We also define a predictive function $f(\cdot)$. Therefore, we define a domain as $\mathcal{D}=\{\mathcal{X},P(\mathbf{X})\}$ and a task as $\mathcal{T}=\{\mathcal{Y},f(\cdot)\}$.

\subsection{Multiple Domains}

Let's define $S$ as the number of domains, where the index of a domain is represented by $s=\{1,...,S\}$. In our MDL approach, we use data from multiple domains $\{\mathcal{D}^s\}^S_{s=1}$ to create a function $p(\cdot)$ that produces a domain-invariant space with encoded features $\mathbf{E} \in \mathbb{R}^{n{\times}{d_e}}$, where $d_e$ is the dimension for the feature space. Ideally, classifiers trained on this domain-invariant space can generalize to all domains. 
\subsection{Missing Values}
In many cases, a feature space from a source domain is incomplete. Given an incomplete feature space $\mathbf{X}$ with missing values, we want to find a function $g(\cdot)$ that can learn the data distribution in the domain and impute values that are as close as possible to the original ones. 

\begin{figure*}[tbp]
\centerline{\includegraphics[width=0.95\linewidth]{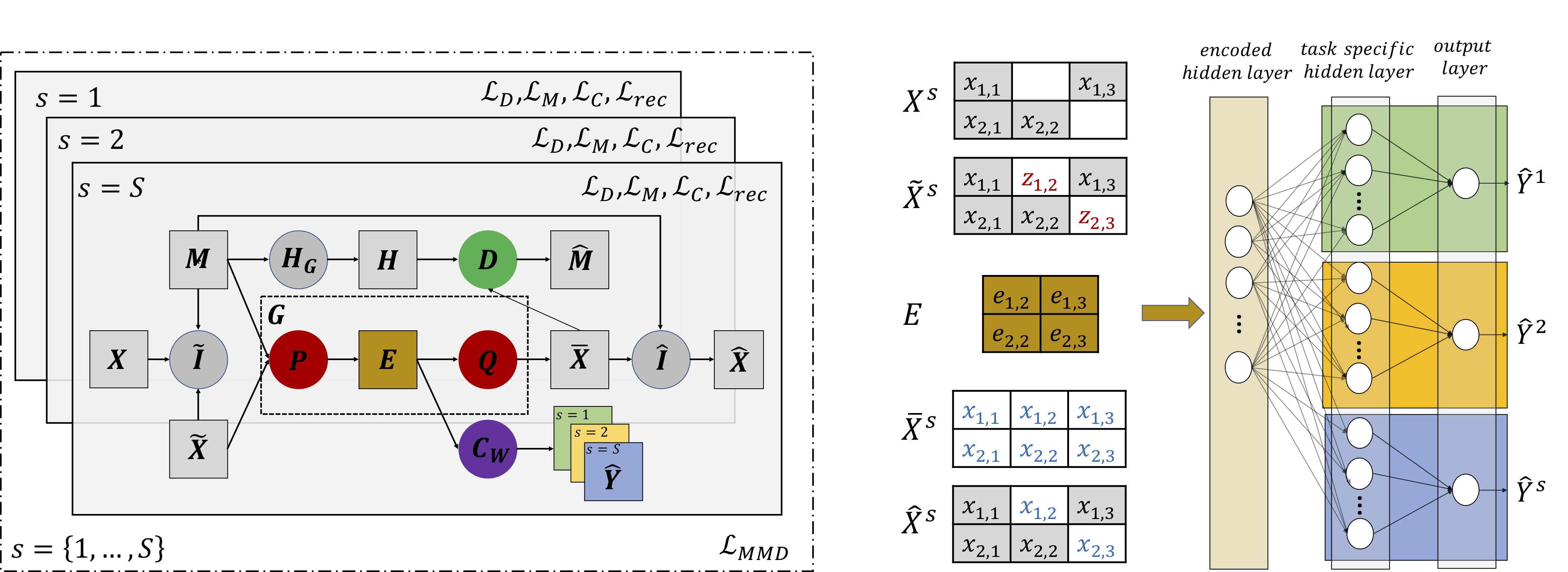}}
\caption{MD2I framework. Left: Overview of the process. Each square represents a dataset. Each circle represents a function that receives and processes input datasets and produces modified output datasets. Components in gray are specific to the current domain, while color components are trained and updated across all domains. Given a domain $s$, a dataset $\mathbf{X^s}$ is fed to the adversarial autoencoder. A generator $G=(P,Q)$ and a discriminator $D$ are trained to learn the representation of the data and produce values for data imputation. At the same time, the encoder component $P$ is trained to generate encoded values $\mathbf{E}$ that are domain-invariant. Finally, the parameters $\mathbf{W}$ of the classifier $C$ are trained using MTL to predicted labels for all domains. For each domain, we calculate the reconstruction, classification and adversarial losses. The MMD loss is calculated after all encoded values are produced. Right: Representation of different datasets along the process and MTL structure taking encoded values as input and producing labels, $\{Y^s\}^S_{s=0}$, as output.}
\label{fig:method}
\end{figure*}

More formally, we can define a new feature space $\tilde{\mathbf{X}} \in \mathbb{R}^{n{\times}d}$ by replacing the missing values with a random noise $z$, not previously observed in $\mathbf{X}$. We also define $\mathbf{M} \in \mathbb{R}^{n{\times}d}$ as a mask vector taking values in $\{0, 1\}^d$ that indicates the components of $\tilde{\mathbf{X}}$ that are observed. This process to create $\tilde{\mathbf{X}}$ and $\mathbf{M}$ is represented by the symbol $\tilde{I}$ in Fig.~\ref{fig:method}. 

For imputation, we want to generate samples according to $P(\mathbf{X} | \tilde{\mathbf{X}} = \tilde{\mathbf{x}}^i)$, for each $i$, to fill in the missing data. Or goal is to approximate  $g(\cdot)$ and $p(\cdot)$ in the same autoencoder structure, achieving MDL and DI in the same framework.
\subsection{Multiple Tasks and Small Datasets}
An additional problem we face on real-world data is datasets with a relatively small sample size $n$. 

Consider that features $\mathbf{X}^s$, and labels, $\mathbf{Y}^s$ for $S$ domains, $s=\{1,...,S\}$, are available and that the sample size, $n^s$ for each domain is relatively too small. This can cause classifiers, $\{f^s(\cdot)\}^S_{s=1}$, to overfit the data when trained individually. 

We aim to train a classifier $f(\cdot)$ that can learn simultaneously from all datasets. Ideally, given a feature input $\mathbf{X^s}$ from a domain $s$, $f(\cdot)$ can correctly predict labels in all domains, $\{\mathbf{Y}^s\}^S_{s=1}$, and the performance for $f(\cdot)$ in each domain is better than using classifiers trained separately. This can be achieved using an MTL framework that seeks to improve the generalization performance of multiple related tasks by learning them jointly. Hence, we aim to combine $p(\cdot), g(\cdot)$ and $f(\cdot)$, so that the encoded feature space and a final classifier are learned simultaneously. 
\section{Methods}
\label{sec:method}
To achieve MDL and DI, we propose an autoencoder using a generator $G=(Q,P)$. $Q$ is an encoder that maps values from an input $\mathbf{X} \in \mathbb{R}^{n \times d}$ to a encoded state $\mathbf{E} \in \mathbb{R}^{d \times d_e}$. $P$ is a decoder that reconstructs the values from $\mathbf{E}$ back to $\mathbf{\bar{X}} \in \mathbb{R}^{n \times d}$. A successful autoencoder is able to produce values for $\mathbf{\bar{X}}$ that are as close as possible to $\mathbf{X}$.

Given $S$ different, but related domains, our goal is to create an autoencoder that can generate the input features for all domains. This can be achieved if $\mathbf{E}$ produced by $G$ is a domain-invariant encoded state (See Fig~\ref{fig:method}).
\subsection{Multi-Domain Learning}
To make hidden codes $\mathbf{E}$ invariant underlying different domains, we first need to define a metric to measure domain similarity. We adopt MMD~\cite{gretton2012kernel}, which finds the largest difference in expectations over functions in the unit ball of a reproducing kernel Hilbert space (RKHS)~\cite{gretton2007kernel}. Hence, we introduce an MMD-based loss term $\mathcal{L}_{mmd}(\mathbf{E}^1,...,\mathbf{E}^S)$ on $\mathbf{E}^s$’s among different domains. Similar to~\cite{li2018domain}, given two domains $\mathbf{E}^s$ and $\mathbf{E}^t$, with number of samples $d^s$ and $d^t$ respectively, the MMD loss is given by:
\begin{equation}
\label{eq:mmd}
\begin{aligned}
MMD(\mathbf{E}^s,\mathbf{E}^t)^2=\norm{ \biggl(
    \frac{1}{d^s}\sum^{d^s}_{i=1}\phi(\mathbf{E}^s_i) - \frac{1}{d^t}\sum^{d^t}_{i=1}\phi(\mathbf{E}^t_i)
    \biggr) }_{\mathcal{H}}^2 \\
= \frac{1}{({d^s})^2}\sum^{d^s}_{i=1}\sum^{d^s}_{j=1}k(\mathbf{E}^{s}_{i},\mathbf{E}^{s}_j)
+\frac{1}{({d^t})^2}\sum^{d^t}_{i=1}\sum^{d^t}_{j=1}k(\mathbf{E}^{t}_{i},\mathbf{E}^{t}_j)\\
-\frac{2}{d^s d^t}\sum^{d^s}_{i=1}\sum^{d^t}_{j=1}k(\mathbf{E}^{s}_{i},\mathbf{E}^{t}_j).
\end{aligned}
\end{equation}

We use a map operation $\phi(\cdot)$ to project the domain into an RKHS $\mathcal{H}$. The work in~\cite{fukumizu2009kernel} shows that if an arbitrary distribution of the features, represented by a kernel embedding technique $k(\cdot,\cdot)$~\cite{smola2007hilbert} is characteristic, then the mapping to the $\mathcal{H}$ is injective. This injectivity indicates that the probability distribution is uniquely represented by an element in RKHS. 

Therefore, we have a kernel function $k(\mathbf{e}^{s}_{i},\mathbf{e}^{t}_{j})=\phi(\mathbf{e}^{s}_{i})\phi(\mathbf{e}^{t}_{j})^\intercal$ induced by $\phi(\cdot)$. For this function, we use the RBF kernel with bandwidth $\sigma$ given by
\begin{equation}
k(\mathbf{e}^{s}_{i},\mathbf{e}^{t}_{j})=\exp(-\frac{1}{2\sigma}\Vert \mathbf{e}^{s}_{i} - \mathbf{e}^{t}_{j}\Vert^2).
\end{equation}

Finally,~\cite{li2018domain} shows it is possible to minimize the upper bound of the distribution variance among domains using
\begin{equation}
\label{eq:mmd_loss}
    \mathcal{L}_{mmd}(\mathbf{E}^1,...,\mathbf{E}^S) = 
    \frac{1}{S^2}\sum_{1\leq i,j\leq S} MMD(\mathbf{E}^i,\mathbf{E}^j).
\end{equation}
\subsection{Adversarial Data Imputation}
To also perform data imputation, we modify our generator $G$ with an adversarial component based on~\cite{yoon2018gain}. More specifically, $\mathbf{X}$ is the original input with missing values. In order to indicate which positions are observed and which are missing, we define a variable $\mathbf{M} \in \mathbb{R}^{n \times d}$, with $m_{i,j}=1$ if  $x_{i,j}$ is observed and $m_{i,j}=0$, otherwise. We then create the random variable $\tilde{\mathbf{X}}$, where missing values, indicated by $\mathbf{M}$, are replaced by noise from a noise variable $\mathbf{Z_m}$.

Given realizations of $\tilde{\mathbf{X}}$, $\mathbf{M}$ and $\mathbf{Z_m}$ as input, $G$ produces a vector of imputations $\bar{\mathbf{X}}$, 
\begin{align} 
\label{eq:xbar}
\bar{\mathbf{X}} &= G(\tilde{\mathbf{X}}, \mathbf{M}, (\mathbf{1 - M}) \odot \mathbf{Z_m}).
\end{align}

To create the final output $\hat{\mathbf{X}}$, we copy the original $\mathbf{X}$ and replace the positions for missing values with values from $\bar{\mathbf{X}}$,
\begin{align} 
\label{eq:xhat}
\hat{\mathbf{X}} &= \mathbf{M} \odot \tilde{\mathbf{X}} + (\mathbf{1} - \mathbf{M}) \odot \bar{\mathbf{X}}.
\end{align}

The process in Eq.~\ref{eq:xhat} is represented by $\hat{I}$ in Fig.~\ref{fig:method}. 

Inspired in~\cite{yoon2018gain}, we introduce a discriminator, $D$, that will be used as an adversary to train $G$. However, instead of predicting if the entire dataset produced by $G$ is fake or real, we use $D$ to predict which values in $\hat{\mathbf{X}}$ are observed and which are imputed. Therefore, a successful $D$ produces values in $\hat{\mathbf{M}}$ that are as close as possible to the real values in $\mathbf{M}$.

We also add a hint mechanism to limit the distributions produced by $G$ that would be optimal for $D$. This hint mechanism is a random variable, $\mathbf{H}$ that depends on $\mathbf{M}$. For each (imputed) sample $(\hat{\mathbf{x}}, \mathbf{m})$, we draw $\mathbf{h}$ according to the distribution $\mathbf{H} | \mathbf{M} = \mathbf{m}$. We pass $\mathbf{h}$ as an additional input to the discriminator so that the $i$-th component of $D(\hat{\mathbf{x}}, \mathbf{h})$ corresponds to the probability that the $i$-th component of $\hat{\mathbf{x}}$ was observed conditional on $\hat{\mathbf{X}} = \hat{\mathbf{x}}$ and $\mathbf{H} = \mathbf{h}$. 

To generate $\mathbf{H}$, we use a random variable $\mathbf{Z_h}$ defined by first sampling $k$ from ${1,...,d}$ uniformly at random, such as
\begin{equation}
    \mathbf{H} = \mathbf{Z_h}\odot \mathbf{M} +0.5(1-\mathbf{Z_h}).
\end{equation}

We can define a minimax problem where we train $D$ to {\em maximize} the probability of correctly predicting $\mathbf{M}$ and train $G$ to {\em minimize} the probability of $D$ predicting $\mathbf{M}$.
\begin{equation} 
\label{eq:minimax}
\begin{split}
\min_{G} \max_{D}  \ &\mathbb{E}_{\hat{\mathbf{X}}, \mathbf{M}, \mathbf{H}}\Big[\mathbf{M}^T \log D(\hat{\mathbf{X}}, \mathbf{H}) \\\nonumber
&\quad+ (\mathbf{1 - M})^T \log\big(\mathbf{1} - D(\hat{\mathbf{X}}, \mathbf{H})\big)\Big],
\end{split}
\end{equation}
where $\log$ is element-wise logarithm and dependence on $G$ is through $\hat{\mathbf{X}}$. Writing $\hat{\mathbf{M}} = D(\hat{\mathbf{X}}, \mathbf{H})$, we can define
\begin{equation} \label{eq:reform}
\min_G\max_D \mathbb{E}\big[\mathcal{L}(\mathbf{M}, \hat{\mathbf{M}})\big],
\end{equation}
where $\mathcal{L}$ is the cross-entropy function defined by
\begin{equation} 
\label{eq:loss}
\mathcal{L}(\mathbf{a},\mathbf{b})=\sum_{i=1}^{d}\Big[a_i\log(b_i)+(1-a_{i})\log(1-b_i)\Big].
\end{equation}
\subsection{Multi-Task Classification}
To incorporate label information into the training process, we define a classifier $C$ with parameters $\mathbf{W} \in \mathbb{R}^{{d_e}\times m}$ that is trained with the encoded values $\mathbf{E}$. By using MTL, we train $C$ to predict the labels $\{\mathbf{Y}^s\}^S_{s=1}$ in each domain in parallel and we use the loss for each task to update $C$ and $G$. 

To introduce sparsity into the model to reduce model complexity, we use a simplification of Lasso in MTL~\cite{zhou2011malsar}. Such regularization controls the sparsity shared among all tasks, assuming that different tasks share the same sparsity parameter. The equation to optimize $C$ is given by
\begin{equation}
\label{eq:loss_mtl}
    \min_\mathbf{W} \sum^S_{s=1}L_C({\mathbf{W}}^T\mathbf{E},\mathbf{Y}^s)
    + \rho_0 \Vert \mathbf{W}\Vert +\rho_{L2} \Vert \mathbf{W}\Vert^2_F,
\end{equation}
where $L_C$ is a standard cross-entropy function, $\mathbf{E}$ are the encoded values produced by $Q$ and $\mathbf{Y}^s$ denotes its corresponding labels in domain $s$. The regularization parameter $\rho_0$ controls sparsity, and $\rho_{L2}$ controls the $l_2$-norm penalty.
\subsection{Training Procedure}
Let's recap that our generator $G$ generates values $\bar{\mathbf{X}}$ for all the positions in $\mathbf{X}$. Therefore, we can use the real and predicted values to calculate the reconstruction error as
\begin{align}
\label{eq:rec_loss}
 \mathcal{L}_{rec}= \sum^{n}_{i=1}\sum^{d}_{j=1}m_{ij}L(\hat{x}_{ij},x_{ij}),
\end{align}
where
\[
 L_{rec}(x_{ij},\hat{x}_{ij})=\begin{cases}
(\hat{x}_{ij}-x_{ij})^{2}, & \text{if \ensuremath{x_{ij}} is continuous},\\
-x_{ij}\log(\hat{x}_{ij}), & \text{if \ensuremath{x_{ij}} is binary}.
\end{cases}
\]

The loss in equation~\ref{eq:rec_loss} ensures that generated values for observed components ($m_{ij} = 1$) are close to those observed. 

We also optimize the generator $G$ to ensure that the imputed values for missing components ($m_{i} = 0$) successfully fool the discriminator using
\begin{align}
\label{eq:m_loss}
 \mathcal{L}_M=-\sum_{i : m_i = 0}(1-m_{i})\log(\hat{m}_{i}),
\end{align}

For the discriminator $D$, we want to ensure that it does not overfit to the hint mechanism. Therefore we only train $D$ to give the outputs that depend on $G$, i.e. $m_i = 0$, using
\begin{align}
\label{eq:discriminator}
\mathcal{L}_D= \sum_{i : m_i = 0} \Big[&m_i\log(\hat{m}_{i})+(1-m_{i})\log(1-\hat{m}_{i})\Big]
\end{align}
$\mathcal{L}_{M}$ is minimized when $D$ fails to correctly reproduce $\mathbf{M}$ and $\mathcal{L}_{rec}$ is minimized when $G$ can produce values close to the observed values. 

The final loss for parameter optimization is given by
\begin{equation}
    \min_{G}\max_D \ \sum_{s=1}^S\Big[
    \mathcal{L}_{rec}
    +\lambda_0\mathcal{L}_{M}
    +\lambda_1\mathcal{L}_{D}
    +\lambda_2\mathcal{L}_{C}\Big]
    +\lambda_3\mathcal{L}_{mmd},
\end{equation}
where hyperparameters $\{\lambda_i\}^3_{i=0}$ can be defined using cross-validation (See Section~\ref{subsec:network_structure} for details on hyperparameters)

The overall algorithm is described in Algorithm \ref{alg:pseudo}. In lines 5, 10 and 14, Algorithm \ref{alg:pseudo} calls the procedure shown in Algorithm \ref{alg:procedure} to generate the necessary variables. In Algorithm \ref{alg:pseudo}, lines 3 to 7 are related to the optimization of $D$. Lines 8 to 12 are the steps to optimize $G$ using the optimized $D$. In lines 13 to 15, the parameters $\mathbf{W}$ for the classifier ${C}$ are updated using the optimized $G$. Lines 17 and 18, guarantee that $G$ is also optimized to create domain-invariant encoded spaces. All optimizations use stochastic gradient descent (SGD).

One can notice that the method can be used in an unsupervised way by skipping the update in lines 13 to 15. This will produce results that are independently of labels in the dataset.
\begin{algorithm}[tb]
\caption{AAE}
\label{alg:procedure}
\textbf{Input}: $\mathbf{X},\mathbf{Z_m},\mathbf{Z_h},G$\\
\textbf{Output}: $\bar{\mathbf{X}},\hat{\mathbf{X}},\mathbf{H},\mathbf{M},\mathbf{E}$
\begin{algorithmic}[1] 
    \STATE $\tilde{\mathbf{X}},\mathbf{M}\leftarrow \tilde{I}(\mathbf{X},\mathbf{Z}_m)$
    \STATE $\bar{\mathbf{X}},\mathbf{E}\leftarrow G(\tilde{\mathbf{X}}, \mathbf{M}, (\mathbf{1 - M}) \odot \mathbf{Z_m})$
    \STATE $\hat{\mathbf{X}} \leftarrow \mathbf{M} \odot \tilde{\mathbf{X}} + (1-\mathbf{M})\odot \bar{\mathbf{X}}$
    \STATE $\mathbf{H} \leftarrow \mathbf{Z}_h\odot \mathbf{M} +0.5(1-\mathbf{Z}^h)$
\end{algorithmic}
\end{algorithm}
\begin{algorithm}[tb]
\caption{MD2I}
\label{alg:pseudo}
\textbf{Input}: $\mathbf{X}=\{\mathbf{X}^1,...,\mathbf{X}^S\}, 
\mathbf{Y}=\{\mathbf{Y}^1,...,\mathbf{Y}^S\}, \mathbf{Z}_m, \mathbf{Z}_h, \\
\rho_1,
\rho_{L2}, \alpha,\{\lambda_1,...,\lambda_4\}$ \\
\textbf{Parameter}: Initialized parameters $G=(Q,P),D,W$\\
\textbf{Output}: $G^*,D^*,W^*$
\begin{algorithmic}[1] 
\WHILE{training loss not converge}
    \FOR{$s=1$ to $S$}
        \STATE draw $B$ batches with $n_b$ samples $\{\mathbf{X}^s_b\}$ from $\mathbf{X}$
        \FOR{$b=1$ to $B$}
            \STATE $\bar{\mathbf{X}^s},\hat{\mathbf{X}^s},\mathbf{H}^s,\mathbf{M}^s,\mathbf{E}^s \leftarrow AAE(\mathbf{X^s},\mathbf{Z_m},\mathbf{Z_h},G)$
            \STATE Update $D$ using SGD and equation~\ref{eq:discriminator}
        \ENDFOR
        \STATE draw $B$ batches with $n_b$ samples $\{\mathbf{X}^s_b\}$ from $\mathbf{X}$
        \FOR{$b=1$ to $B$}
            \STATE $\bar{\mathbf{X}^s},\hat{\mathbf{X}^s},\mathbf{H}^s,\mathbf{M}^s,\mathbf{E}^s \leftarrow AAE(\mathbf{X^s},\mathbf{Z_m},\mathbf{Z_h},G)$
            \STATE Update $G$ using SGD and equations~\ref{eq:rec_loss} and ~\ref{eq:m_loss}.
        \ENDFOR
        \STATE draw 1 batch with $n_b$ samples $\{\mathbf{X}^s_b\}$ from $\mathbf{X}$
        \STATE $\bar{\mathbf{X}^s},\hat{\mathbf{X}^s},\mathbf{H}^s,\mathbf{M}^s,\mathbf{E}^s \leftarrow AAE(\mathbf{X^s},\mathbf{Z_m},\mathbf{Z_h},G)$
        \STATE Update $W$ and $G$ using SGD and equation~\ref{eq:loss_mtl}
        
    \ENDFOR
    \STATE Calculate loss $\mathcal{L}_{mmd}$ using equation ~\ref{eq:mmd_loss}
    \STATE Update $G$ using SGD and $\mathcal{L}_{mmd}$
    
\ENDWHILE
\end{algorithmic}
\end{algorithm}

\section{Experiments}
\label{sec:experiments}
We perform experiments on real-world problems and compare our framework with other methods in distinct settings.
\subsection{DG-DI in Image Recognition with unstructured data}
We first apply our method to image recognition using the MNIST~\cite{lecun1998gradient} dataset. To create the dataset for DG, we used the method proposed in~\cite{ghifary2015domain} by producing digit images from six different angles. We use 1,000 digit images of ten classes (with 100 images per class) to represent the basic view. We denote the digit images with 0$^{\circ}$ by $I_0$, and rotate the images in a counter-clockwise direction by 15$^{\circ}$, 30$^{\circ}$, 45$^{\circ}$, 60$^{\circ}$ and 75$^{\circ}$. 

To create the dataset for DG-DI, we remove a square patch of the size of 13 $\times$ 13 from all images (each image is 28 $\times$ 28 = 784 pixels). The location of the patch was uniformly sampled for each image. This creates a missing completely at random (MCAR) case, where there is no dependency on any of the variables. We also perform normalization to the range [0, 1] on pixels and feed the final values to the methods. 

For validation, we apply the leave-one-domain-out strategy, using one of the domains as target and training the model in all other domains as a source. We apply the missing mask in both source and target. Finally, we evaluate the methods in terms of classification using the accuracy score. We repeat the experiments for 50 times and report the averaged results.
\subsection{MTL-DI in Regression with structured data}
In this setting, we use the school dataset\footnote{Available at http://multilevel.ioe.ac.uk/intro/datasets.html}, which consists of exam grades from 15,362 students in 139 schools. This is a structured dataset with categorical and numerical features, where categorical attributes have been expressed as binary attributes using one-hot encoding. Our goal is to predict each student’s exam grade, hence, the school is the task and exam score is the output, creating an MTL setting. 

For MTL-DI, we introduce missingness by appending a uniform random vector $\mathbf{r}$ with values between 0 and 1. Each observation in $r$ corresponds to an observation in the dataset. We then randomly sample two attributes $x_1$ and $x_2$ from the dataset and calculate their median $mx_1$ and $mx_2$. Finally, we pick half of the attributes, at random, to have missing values where $r_i\leq t, i\in 1:n$ and $(x_1\leq mx_1$ or $x_2 \geq mx_2)$, where $n$ is the number of observations and $t$ is a threshold, that we define as 30\%. This creates a missing at random (MAR) case, where the missingness depends on the observed variables.

For validation, we perform 50 experiments with a 50/50\% training/test split in each one. Methods are compared in terms of Root Mean Squared Error (RMSE) on the test set.
\subsection{MDMTL-DI in Classification with mixed data}
We also evaluate our method in a real-world problem with mixed data, i.e. structured data with continuous and categorical features, and text data. Our information comes from 3 different companies that use selection processes with a similar structure. 

For privacy requirements, we refer to the companies with indexes, such that $C_1$ refers to Company 1. $C_1$ is an organization that selects students to receive fellowships. $C_2$ is a big retail company that selects students to work in the company after graduation. $C_3$ is a governmental agency that selects students for positions in public administration. Table~\ref{tab:process} presents the description for each stage in each process. All processes happen yearly and we have data from two years.

Each stage in the process contains its own set of variables. The data collected in each process is very similar in stages such as \textit{Demographics} (Demo) and \textit{Education} (Edu), both in the form of content and structure. For stages with open questions such as \textit{Video Submission} and \textit{Experience} (Exp), each process has its own specific questions.
\begin{table}[tbp]
\caption{Stages in the selection processes}
\label{tab:process}
\scriptsize
\begin{tabularx}{\columnwidth}{|>{\hsize=.1\hsize\columnwidth=\hsize}X|>{\hsize=.40\hsize\columnwidth=\hsize}X|>{\hsize=.25\hsize\columnwidth=\hsize}X|>{\hsize=.25\hsize\columnwidth=\hsize}X|}

\hline
\heading{Stages} & \heading{$C_1$} & \heading{$C_2$}  & \heading{$C_3$}  \\ \hline
Demo & Provide country, city, age. & Same as $C_1$. & Same as $C_1$. \\ \hline
Edu & Provide university, major, minor and extra activities. & Same as $C_1$. & Same as $C_1$. \\ \hline
Profile Test & Online tests to measure profile characteristics such as ambition and interests. & Online tests to measure big 5 characteristics & Same as $C_1$. \\ \hline
Exp & Write about professional experience (Open format). & Same as $C_1$ & Same as $C_1$. \\ \hline
Logic Test & Perform online tests to map levels in problem-solving involving logic puzzles. & Same goal as $C_1$ but with specific test for $C_2$ & Same goal as $C_1$ but with specific test for $C_3$ \\ \hline
Video Submission & 2-min, explaining why they deserve the fellowship. & 5-min, making a case to be selected for the position & 1-min, explaining a problem that motivates the applicant \\ \hline
Video Evaluation & Applicants are evaluated based on their entire profile. & Same as $C_1$ but different criteria. & Same as $C_1$ but different criteria. \\ \hline
\end{tabularx}
\end{table}
\subsubsection{Feature preparation and missing data}
Categorical variables are converted to numerical using a one-hot encoder. In stages such as video, the speech is extracted and the data is used as a text. To convert text data to numerical values, we create word embeddings using Word2Vec~\cite{mikolov2013efficient}, where we assign high-dimensional vectors to words in a text corpus but preserving their syntactic and semantic relationships.

Approximately, 30\% of the information is missing across all datasets. The source of the missing data is not available, but from different observations, we estimate that it has cases of both MCAR and MAR.
\subsubsection{Validation and performance metrics for selection}
Our goal is to predict which applicants will be selected in the \textit{Video Evaluation} stage. Although the real evaluation in each process happens independently, we believe the datasets are related due to the similar structure of the features and the common focus on undergraduate students in their last year. 
 
We perform longitudinal experiments, in which we use a $year_1$ as a training and test set and $year_2$ as a validation set. We also use the standard cross-validation (CV) setting by using data from a specific year and splitting it into train, test, and validation. We repeat this process 3 times, one for each year and one for a dataset with both years combined. 

We use a 10-fold CV in each experiment with a total of 50 runs. Furthermore, we compare the models in terms of F1-score for the positive class, which balances precision and recall with a focus on the selected applicants.
\subsection{Benchmark Methods}
In our baseline method (B1), we train an individual classifier for each domain. The second baseline (B2) uses the aggregation of all domains into one and trains one individual classifier for all domains. For missing data, we perform mean-imputation, which imputes the mean of each continuous attribute and the mode of each discrete attribute.

For comparison in DI, we repeat the baselines (B1, B2) but use GAIN~\cite{yoon2018gain} to perform data imputation.
Next, we compare with other MDL methods, using: D-MTAE~\cite{ghifary2015domain}, MMD-AAE~\cite{li2018domain} and CCSA~\cite{motiian2017unified}.
At last, we use methods that address MTL such as DMTRL~\cite{yang2016deep}, and UMDMTL (Unified MD and MTL)~\cite{yang2014unified} that addresses both MTL and MDL. 

For all methods, we use network specification, hyperparameters, and initialization as reported in the respective papers. For prediction in D-MTAE, Baseline and DI methods, we use an NN with 2 hidden layers with cross-entropy loss for classification and RMSE for regression.

\begin{table}[tbp]
\caption{Benchmark Methods and their components.}
\label{tab:components}
\scriptsize
\centering
\begin{tabular}{|c|c|c|c|c|c|}
\hline
Method & Group & Noise & Labels & MTL & Domain \\ \hline
B1 & Baseline &  & x &  &  \\ \hline
B2 & Baseline &  & x &  &  \\ \hline
B1-DI & DI & x & x &  &  \\ \hline
B2-DI & DI & x & x &  &  \\ \hline
CCSA & MDL &  & x &  & x \\ \hline
D-MTAE & MDL & x &  &  & x \\ \hline
MMD-AAE & MDL &  & x &  & x \\ \hline
DMTRL & MTL &  & x & x &  \\ \hline
UMDMTL & MDMTL &  & x & x & x \\ \hline
MD2I-U & Ours & x &  &  & x \\ \hline
MD2I-S & Ours & x & x & x & x \\ \hline
\end{tabular}
\end{table}
In Table~\ref{tab:components}, we present all methods, the components that they address and their group. \textit{Noise} and \textit{Domain} are marked if the method directly addresses DI and MDL, respectively. In the training process, \textit{Labels} is marked if labels are used and \textit{MTL} is marked if there is an MTL component.
\subsection{Network Structure}
\label{subsec:network_structure}
We use a single hidden layer for the autoencoder with dimension $d_e=\frac{d}{log(d)}$. This layer is the input for a prediction layer with 10 neurons for each task with single-output (binary or regression). If the task is categorical with $d_y$ classes, the number of neurons is $\max(10,d_y)$. We use ReLU as the non-linear activation and SGD optimizer with batch size$=$32, learning rate$=$0.01 and momentum$=$0.9. We obtain consistent results across all experiments using $\sigma=10$ in the RBF kernel,
$\{\lambda_i\}^3_{i=0}$ set to 1,0.1,0.1 and 2, respectively, for the final loss and, $\rho_0=1$ and $\rho_{L2}=0.1$ for the regularization in the MTL classifier. We test our approach using two configurations: unsupervised (MD2I-U) and supervised (MD2I-S), which uses labels and MTL in the training process. MD2I-U is used to generate the encoded values and a final 2-layer NN classifier is used for prediction.
\section{Results and Ablation Study}
\label{sec:results}
Here, we report the results for all settings and discuss our main findings. We also highlight the importance of the components shown in Table~\ref{tab:components} and their impact on performance.
\subsection{DG-DI for Image Recognition}
\begin{table}[tbp]
\caption{Results in DG-DI setting for image recognition}
\label{tab:image_results}
\scriptsize
\centering
\begin{tabular}{l|l|l|l|l|l|l|l|}
\cline{2-8}
 & $I_0$ & $I_{15}$ & $I_{30}$ & $I_{45}$ & $I_{60}$ & $I_{75}$ & \textbf{Avg} \\ \hline
\multicolumn{1}{|l|}{MD2I-S} & \textbf{80.8} & 89.5 & \textbf{91.3} & \textbf{82.7} & \textbf{93.6} & 74.5 & \textbf{85.4} \\ \hline
\multicolumn{1}{|l|}{MD2I-U} & \textbf{79.9} & \textbf{90.3} & 91.2 & \textbf{82.4} & \textbf{92.4} & 74.7 & \textbf{85.2} \\ \hline
\multicolumn{1}{|l|}{D-MTAE} & 79.6 & 89.4 & \textbf{91.8} & 80.2 & 90.7 & \textbf{77.1} & 84.8 \\ \hline
\multicolumn{1}{|l|}{MMD-AAE} & 77.2 & \textbf{92.8} & 87.9 & 72.2 & 90.0 & \textbf{75.0} & 82.5 \\ \hline
\multicolumn{1}{|l|}{CCSA} & 72.1 & 86.8 & 85.7 & 76.5 & 89.1 & 70.9 & 80.2 \\ \hline
\multicolumn{1}{|l|}{B2-DI} & 66.4 & 84.3 & 83.8 & 76.2 & 84.9 & 69.2 & 77.5 \\ \hline
\multicolumn{1}{|l|}{B2} & 48.2 & 69.1 & 65.4 & 55.6 & 65.2 & 51.0 & 59.1 \\ \hline
\end{tabular}
\end{table}
Table~\ref{tab:image_results} show the results obtained for the image recognition task. By comparing B2 and B2-DI, we see that the DI component is important as it improves performance in 31\%. For DG methods, all can improve upon baselines. However, CCSA has an inferior performance compared to D-MTAE and MMD-AAE. This can be explained by the autoencoder structure that can generalize well across domains while being robust to incomplete data. In fact, D-MTAE is designed to use noise to improve robustness, making its performance better than MMD-AEE, which does not directly handle missing data. 

The best results are achieved by our method in both versions (MD2I-S, MD2I-U), showing that it can produce domain-invariant features and better learn the distribution of the data. 

In this setting, including the labels during training does not significantly improve the performance in our methods and does not help MMD-AAE to outperform D-MTAE. In general, we see that methods that address the \textit{Noise} and \textit{Domain} components have better performance.  
\subsection{MTL-DI for Grade Prediction}
\begin{figure}[tbp]
\centerline{\includegraphics[width=\columnwidth]{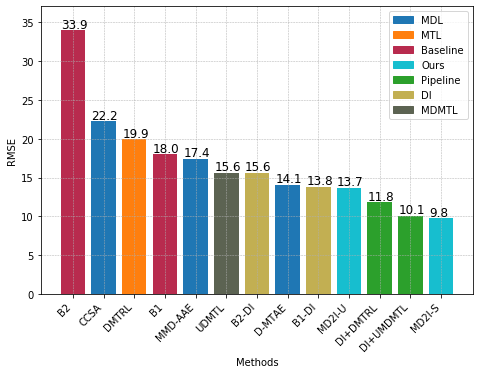}}
\caption{Results in MTL-DI setting for grade prediction.}
\label{fig:mtl_result}
\end{figure}

For the results in the MTL-DI setting shown in Figure~\ref{fig:mtl_result}, we first observe that B1 significantly outperforms B2, which indicates that using all domains together can cause negative transfer. However, the influence of DI seems to be higher than MDL, as B1-DI and B2-DI outperform MDL methods, such as CCSA and MMD-AAE. By comparing D-MTAE (which handles DI and MDL but no labels) and B1-DI, we see that using labels in training improves performance.

DMTRL and UMDMTL alone can't beat the baselines, as they are affected by the missing data. To improve them, we perform experiments with a pipeline. First, we use the DI method to generate a complete dataset and then the MTL methods to predict outputs. These new pipelines achieve better scores, with both DI+UMDMTL and DI+DMTLR outscoring MD2I-U. We also implement MDL methods using a pipeline, but the results were similar to the best baseline. 

The best results are achieved by our supervised method (MD2I-S), especially due to the robustness to incomplete datasets and the inclusion of label information in training using an MTL approach. Therefore, for this setting, we see that \textit{Labels} and \textit{Noise} have higher impact while \textit{Domain} has lower impact in performance.
\subsection{MDMTL-DI for Classification in Selection Process}
\begin{table}[tbp]
\caption{Results in MDMTL-DI setting for Classification}
\label{tab:selection_results}
\centering
\scriptsize
\begin{tabular}{|c|c|c|c|c|c|}
\hline
Method & Group & $C_1$ & $C_2$ & $C_3$ & Avg \\ \hline
MD2I-S & Ours & \textbf{83.2} & \textbf{69.7} & \textbf{63.7} & \textbf{72.2} \\ \hline
DI+UMDMTL & Pipeline & \textbf{79.3} & \textbf{67.1} & \textbf{61.2} & \textbf{69.2} \\ \hline
MD2I-U & Ours & \textbf{78.5} & \textbf{60.4} & \textbf{54.7} & \textbf{64.5} \\ \hline
D-MTAE & MDL & 69.9 & 55.1 & 52.4 & 59.1 \\ \hline
DI+DMTRL & Pipeline & 67.4 & 57.6 & 52.3 & 59.1 \\ \hline
UMDMTL & MDMTL & 71.6 & 57.7 & 45.7 & 58.3 \\ \hline
MMD-AAE & MDL & 72.9 & 52.2 & 48.6 & 57.9 \\ \hline
CCSA & MDL & 53.7 & 46.5 & 37.5 & 45.9 \\ \hline
DMTRL & MTL & 42.5 & 33.4 & 36.4 & 37.4 \\ \hline
B2-DI & DI & 43.2 & 25.0 & 38.5 & 35.6 \\ \hline
B1-DI & DI & 44.6 & 18.0 & 37.3 & 33.3 \\ \hline
B2 & Baseline & 41.3 & 18.5 & 33.5 & 31.1 \\ \hline
B1 & Baseline & 42.4 & 15.5 & 31.2 & 29.7 \\ \hline
\end{tabular}
\end{table}
Table~\ref{tab:selection_results} shows the results for the selection processes. In terms of the rank of the methods, we can see that the bottom 6 (from CCSA down) and the top 3 are consistent across all companies. The positions from 4 to 7 change for each company, showing that none of the methods (MMD-AAE, UMDMTL, D-MTAE, and DI+DMTRL) in this middle tier have a general superior performance over the others.

When comparing methods in the average column, the worst ones are the baselines, which shows that individual classifiers for each process achieve inferior results. In this setting, DI does not have a high impact as previously, as observed in the marginal gain comparing B1 and B1-DI. In fact, the component \textit{Domain} has a higher impact in this setting, as MDL methods can significantly outperform DI methods.

By having MDL and MTL, UMDMTL can beat DI methods and stay in the middle tier. However, when combined with DI using a pipeline, it consistently achieves the second top score, repeating the results from the MTL-DI setting. 

MD2I-U is the third-best, beating all other methods but falling short to the top due to not using label information in training. We show that our method can produce domain-invariant encoded spaces and address data imputation jointly. By performing it without labels, we suspect that the features generated from this method can be better generalized to unseen domains where predictions are not available. 

Nonetheless, when labels are available, our supervised, MD2I-S, is proven to be effective as it achieves the best score in all settings. These results show the importance of learning all components together in the same framework.

\section{Conclusion}
\label{sec:conclusion}
We presented a method to combine MDL, DI, and MTL in a single framework. Our loss function optimizes an adversarial autoencoder to jointly produce domain-invariant encoded spaces and learn the data distribution for data imputation. In the unsupervised version (MD2I-U), the encoded and generated features can be used for different downstream tasks. In the supervised version, we add an MTL component for predictions in all domains and to learn from different tasks to improve a single classifier. By combining small datasets from different but related domains, our method can perform better than any classifier trained separately. Through many experiments in different settings including structured, unstructured and mixed data, our method outperformed other state-of-art methods, by successfully combining important components in the same structure. For future work, we want to investigate the features generated and how they can be applied for clustering and classification in unlabelled datasets in different domains.

\bibliographystyle{IEEEtran}
\bibliography{main}

\end{document}